\documentclass[lettersize,journal]{IEEEtran}
\usepackage{amsmath,amsfonts}
\usepackage{algorithmic}
\usepackage{algorithm}
\usepackage{array}
\usepackage[caption=false,font=normalsize,labelfont=sf,textfont=sf]{subfig}
\usepackage{textcomp}
\usepackage{stfloats}
\usepackage{url}
\usepackage{verbatim}
\usepackage{graphicx}
\usepackage{cite}
\usepackage{orcidlink}
\usepackage{hyperref}
\hypersetup{hypertex=true,
            colorlinks=true,
            linkcolor=blue,
            anchorcolor=blue,
            citecolor=blue,
            linktocpage}
\usepackage{multirow}
\usepackage{makecell}
\usepackage{setspace}
\usepackage{booktabs}
\usepackage{amssymb}
\begin{document}

\title{QdaVPR: A novel query-based domain-agnostic model for visual place recognition}

\author{Shanshan Wan$^{\orcidlink{0009-0007-8497-1582}}$, Lai Kang, Yingmei Wei, Tianrui Shen$^{\orcidlink{0000-0001-5300-4396}}$, Haixuan Wang, and Chao Zuo
        % <-this % stops a space
\thanks{Corresponding authors:
Lai Kang and Yingmei Wei.}% <-this % stops a space
\thanks{Shanshan Wan, Lai Kang, Yingmei Wei, Tianrui Shen, Haixuan Wang and Chao Zuo are with the College of Systems Engineering and the Laboratory for Big Data and Decision, National University of Defense Technology, Changsha, Hunan 410073, China (e-mail: wanshanshan16@nudt.edu.cn; kanglai@nudt.edu.cn; weiyingmei@nudt.edu.cn; shentianrui@nudt.edu.cn; wang77@nudt.edu.cn; zuochao2002@nudt.edu.cn).}}

\markboth{\tiny{This work has been submitted to the IEEE for possible publication. Copyright may be transferred without notice, after which this version may no longer be accessible.}}%
{Shell \MakeLowercase{\textit{et al.}}: A Sample Article Using IEEEtran.cls for IEEE Journals}

\IEEEpubid{0000--0000/00\$00.00~\copyright~2021 IEEE}

\maketitle

\begin{abstract}
Visual place recognition (VPR) aiming at predicting the location of an image based solely on its visual features is a fundamental task in robotics and autonomous systems. Domain variation remains one of the main challenges in VPR and is relatively unexplored. Existing VPR models attempt to achieve domain agnosticism either by training on large-scale datasets that inherently contain some domain variations, or by being specifically adapted to particular target domains. In practice, the former lacks explicit domain supervision, while the latter generalizes poorly to unseen domain shifts. This paper proposes a novel query-based domain-agnostic VPR model called QdaVPR. First, a dual-level adversarial learning framework is designed to encourage domain invariance for both the query features forming the global descriptor and the image features from which these query features are derived. Then, a triplet supervision based on query combinations is designed to enhance the discriminative power of the global descriptors. To support the learning process, we augment a large-scale VPR dataset using style transfer methods, generating various synthetic domains with corresponding domain labels as auxiliary supervision. Extensive experiments show that QdaVPR achieves state-of-the-art performance on multiple VPR benchmarks with significant domain variations. Specifically, it attains the best Recall@1 and Recall@10 on nearly all test scenarios: 93.5\%/98.6\% on Nordland (seasonal changes), 97.5\%/99.0\% on Tokyo24/7 (day-night transitions), and the highest Recall@1 across almost all weather conditions on the SVOX dataset. Our code will be released at \href{https://github.com/shuimushan/QdaVPR}{https://github.com/shuimushan/QdaVPR}.
\end{abstract}


\section{Introduction}
\IEEEPARstart{V}{isual} place recognition (VPR) aims to predict the location of a current observation based on stored geotagged historical observations \cite{netvlad}. It is typically achieved by converting images into comparable descriptors and performing a k-nearest neighbor search between the query descriptor and the database descriptors. As a fundamental capability for robot state estimation, VPR is widely used in applications such as mobile robot localization \cite{application1,application2} and autonomous driving \cite{autonomous}, among other areas.

VPR faces three primary challenges: viewpoint change, perceptual aliasing, and domain variation \cite{survey}. While viewpoint robustness has been extensively studied through viewpoint-aware data sampling strategies \cite{eigenplaces,dsformer,CliqueMining} or partial retrieval technique \cite{SegVLAD}, and perceptual aliasing has been addressed via uncertainty estimation \cite{uncertainty1,uncertainty2,uncertainty3}, domain variation remains relatively underexplored. 

Existing approaches to domain-agnostic VPR roughly fall into two categories. The first trains models on large-scale datasets (e.g., GSV-cities \cite{gsv-cities}) that inherently contain some domain diversity, but lacks explicit domain supervision to guide invariant feature learning. The second employs domain adaptation techniques \cite{domain1,domain2,adageo,daco,CerfeVPR,domain_latest} tailored to specific target domains, yet these methods generalize poorly to unseen domain shifts and often require additional generative models \cite{cyclegan} or target-domain data during training.

\begin{figure}[!t]
\centering
\includegraphics[width=\linewidth]{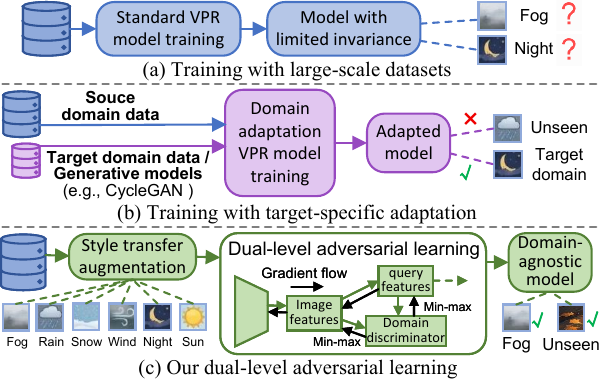}
\caption{Overview of the domain-agnostic VPR models. (a)  Most models acquire some domain agnosticism through training on large-scale datasets.  (b) Some VPR models are specially designed to be robust to target domains but generalize poorly to unseen domains. (c) Our model achieves domain agnosticism via a novel dual-level adversarial learning framework, which makes it more robust to VPR tasks under domain shift.} 
\label{fig0}
\end{figure}

\IEEEpubidadjcol

To address these limitations, we propose QdaVPR, a novel query-based domain-agnostic VPR model built upon the Bag-of-Queries (BoQ) architecture \cite{BoQ}. The key differences between our model and existing domain-agnostic VPR models are illustrated in Fig. \ref{fig0}. BoQ introduces learnable queries that probe image features via cross-attention to aggregate robust global descriptors, whose query features obtained from each BoQ layer are concatenated and weighted to form the final global descriptor. As each query feature encapsulates specific information pertaining to the entire image, they can be directly employed for domain-agnostic supervision. To leverage the large-scale VPR dataset and perform domain-guided supervision for producing a domain-agnostic model, we augment the GSV-cities dataset using a style transfer library \cite{imgaug} to generate six synthetic domains (fog, rain, snow, wind, night, sun) with corresponding domain labels. Then, two novel mechanisms are proposed to enforce domain invariance and enhance discriminability. Specifically, a dual-level adversarial learning framework is introduced, which applies gradient reversal and domain classification to both the query features and the underlying image features, encouraging the model to discard domain-specific information. In addition, the global descriptor in QdaVPR is formed by a fixed set of weighted combinations of the query features, each aggregating domain-agnostic information from all query features. To enhance the discriminative power of these invariant features, a query-combination-based triplet loss is proposed, focusing supervision on reliable query combinations. These mechanisms enable QdaVPR to generalize effectively across unseen domains without introducing additional computational overhead at inference.

The main contributions of our work include:
\begin{itemize}
\item  A dual-level adversarial learning framework that enforces domain invariance at both the query feature and image feature levels. Since query features are derived from image features, this dual-level design enables mutual reinforcement, making both levels more domain-agnostic.
\item  A query-combination-based triplet supervision strategy to mine discriminative descriptors. A triplet loss is applied between each reliable anchor combination and the corresponding combinations from the easiest positive and hard negatives, focusing supervision on the most reliable and challenging components.
\item  Achieving SOTA performance on multiple VPR benchmarks with cross-domain variations. QdaVPR achieves the best Recall@1 and Recall@10 on nearly all test scenarios on the Nordland datasets \cite{nordland} (for seasonal changes), Tokyo24/7 dataset \cite{tokyo} (for day-night variation) and the SVOX datasets \cite{adageo} (for weather and illumination changes). 
\end{itemize}

\section{Related Works}
This section reviews three relevant research areas. We first outline the landscape of domain-agnostic VPR, noting its reliance on large-scale training or explicit robustness mechanisms. Next, we present the related task of domain adaptation in retrieval-based visual localization. Finally, we discuss domain generalization strategies from the broader literature, identifying a gap that motivates our dual-level adversarial learning framework.
\subsection{Domain-agnostic VPR}
VPR can be formulated as an image retrieval task across different geographical locations \cite{netvlad}. Typically, descriptors for database images are first extracted and stored by passing them through a VPR model. For a given inquiry image, its descriptor is obtained using the same model, and the corresponding place is retrieved by performing a k-nearest neighbor (k-NN) search within the database. Based on their retrieval pipeline, VPR methods can be broadly categorized into one-stage and two-stage approaches. One-stage methods directly rely on global descriptors for retrieval, utilizing either hand-crafted local features \cite{sift} aggregated via non-trainable techniques like VLAD \cite{vlad} or, more recently, trainable backbones and aggregation layers \cite{netvlad,sfxl,eigenplaces,salad,BoQ,SciceVPR,djist,ImAge}. Two-stage methods refine initial retrieval results by incorporating a second stage of finer matching, typically using local descriptors on the top-ranked candidates to re-rank them, which improves precision at the cost of additional computation \cite{selavpr,FoL,structvpr++}. While these categories define the general landscape of VPR, the focus of this work is on enhancing model robustness to domain variations, a critical aspect for real-world deployment.

\textbf{Domain-agnostic VPR.} Most VPR models \cite{sfxl,eigenplaces,CliqueMining,dsformer,SciceVPR,djist,gsv-cities,salad,BoQ,edtformer,selavpr,structvpr++,FoL,CerfeVPR,ImAge} acquire a degree of domain invariance by being trained on large-scale datasets like SF-XL \cite{sfxl}, GSV-cities \cite{gsv-cities}, and MSLS \cite{msls}, which contain images under varied conditions but lack explicit domain labels. In contrast, some models are specifically designed to resist certain domain changes \cite{domain1,domain2,adageo,daco,CerfeVPR,domain_latest}. For instance, Qin et al. \cite{domain1} proposed a season-robust VPR model by adversarially disentangling season-invariant content features from season-specific appearance features. Wang et al. \cite{domain2} targeted age-invariant VPR by employing a multi-kernel maximum mean discrepancy loss for domain distribution alignment. AdAGeo \cite{adageo} performed adversarial domain adaptation within a few-shot learning framework for robustness to a specific target domain. Daco \cite{daco} integrated domain translation with contrastive learning to enhance robustness to a particular target domain. Another work by Wang et al. \cite{domain_latest} addressed fine-grained domain adaptation, yet still for a predefined target domain. 

Most recently, CerfeVPR \cite{CerfeVPR} employs CycleGAN \cite{cyclegan} to generate domain variants of input images, processing both original and transformed images through a transformer to produce domain-invariant features. However, this approach introduces an auxiliary generative model and incurs considerable computational overhead during both training and inference, as both original and generated images must be processed. In contrast, our method pre-generates domain variants using a style transfer library \cite{imgaug} and uses them solely as supervisory signals for adversarial learning. The adversarial modules are active only during training and discarded at inference, enabling efficient deployment without extra computational cost.

\textbf{Domain adaptation in retrieval-based visual localization.} The task of domain adaptation within retrieval-based visual localization \cite{DASGIL,Double-Domain} is closely related to domain-agnostic VPR. For instance, DASGIL \cite{DASGIL} addresses domain shift by utilizing synthetic virtual domain data (depth and segmentation) for training, thereby enabling adaptation from virtual to real environments. More recently, Ge et al. \cite{Double-Domain} proposed a dual-domain adaptation framework that explicitly incorporates semantic information, achieving SOTA performance among retrieval-based visual localization methods. While both retrieval-based localization and VPR rely on global descriptor retrieval for localization, VPR imposes a coarser requirement, as it does not demand precise viewpoint alignment between query and database images. In contrast, retrieval-based visual localization typically targets high-precision pose estimation, often within thresholds such as 5 meters and 10 degrees. Consequently, the comparative analysis in this work will focus exclusively on VPR models.

\begin{figure*}[!t]
\centering
\includegraphics[width=0.8\linewidth]{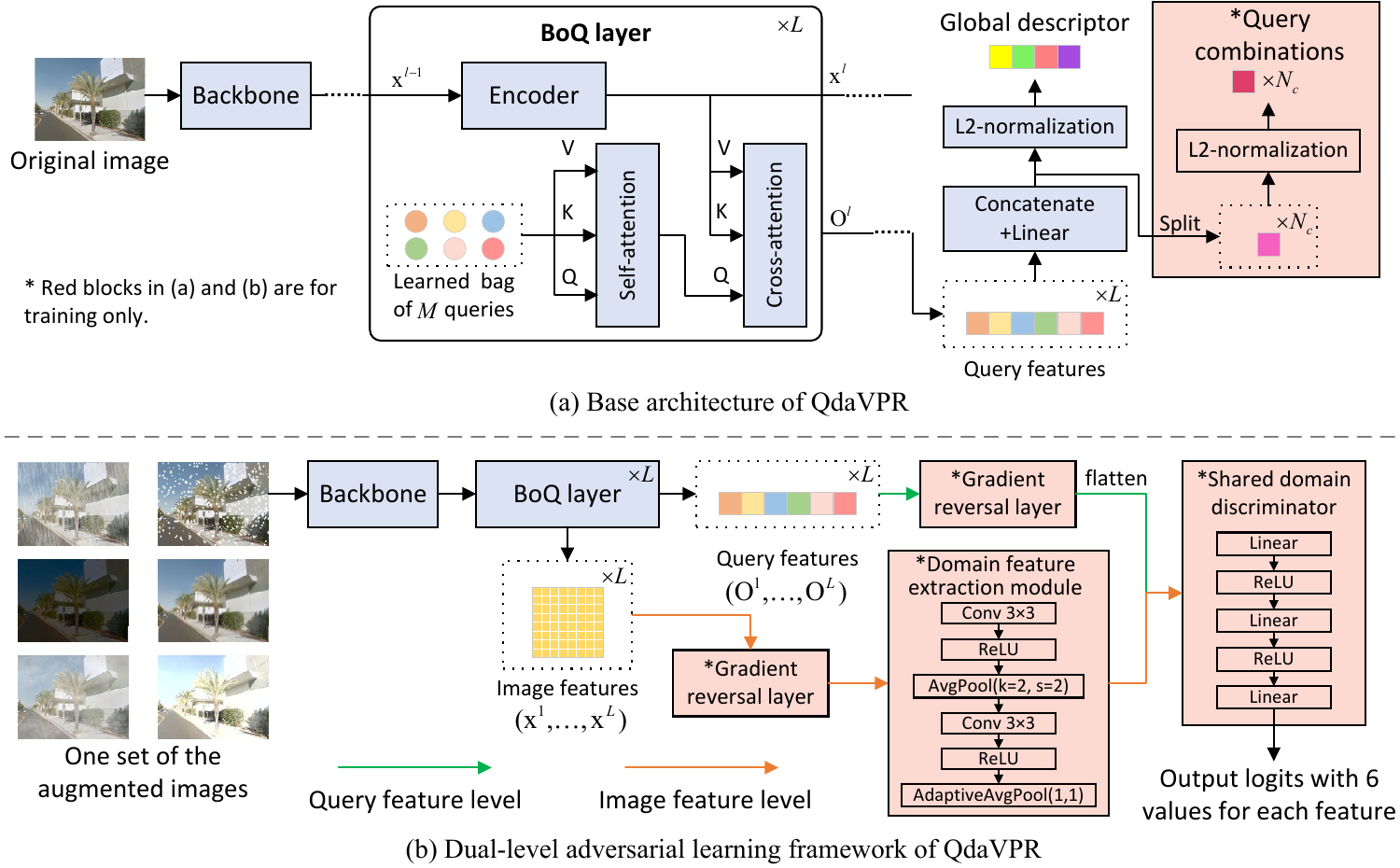}
\caption{Overview of the QdaVPR model. (a) The base architecture. Regardless of whether the input image is from the original GSV-cities or the generated six-domain synthetic GSV-cities datasets, this architecture outputs a global descriptor and \(N_c\) query combinations. (b) The framework for dual-level adversarial learning. Each of the \(L \times M\) query features and the \(L\) domain features (extracted from the \(L\) image features) is fed into a domain discriminator, producing output logits with six values. These values predict the specific domain of the input image. Note that output logits are generated only when the input image is from the generated six-domain dataset; otherwise, the outputs are as shown in (a). During inference, the red blocks are discarded, and only the global descriptor is output. The channel dimension is omitted from the figure for clarity. See text for details.}
\label{fig1}
\end{figure*}

\subsection{Domain generalization}

The domain-agnostic VPR methods discussed above typically address domain shifts by adapting to specific target domains, assuming prior knowledge or accessibility of the target domain during training. In real-world deployments, however, the target domain is often unknown, requiring models that can generalize across arbitrary domains without adaptation. This aligns with the goal of domain generalization, which aims to learn a model from one or multiple source domains such that it performs well on any unseen target domain \cite{domain_generalization_survey}. Unlike domain adaptation, domain generalization does not assume access to target domain data during training.

Various strategies have been developed to achieve domain generalization. Domain alignment methods \cite{domain_alignment1,domain_alignment2,domain_alignment3} seek to learn domain-invariant representations by minimizing distribution differences (e.g., by using adversarial learning to confuse a domain discriminator across source domains \cite{domain_alignment1,domain_alignment2}). Data augmentation methods \cite{data_augment3,data_augment4} actively increase the diversity of training data to cover potential shifts (e.g., by using style transfer models \cite{data_augment5} to map images from one source domain to another). Disentangled representation learning \cite{domain1,Disentangled_representation_learning} allows part of the network to be domain-specific, often through decomposing the model into a domain-specific component and a domain-agnostic component \cite{domain1}. Other approaches leverage meta-learning, ensemble learning, self-supervised learning, regularization strategies, or reinforcement learning. For further details, we refer readers to \cite{domain_generalization_survey}.

Most VPR models apply basic data augmentation techniques \cite{randomaug} during training, but rarely incorporate other domain generalization strategies. Recently, Wang et al. \cite{cross_view} proposed a self-adaptive network for aerial-view geo-localization that uses the same style transfer library \cite{imgaug} for augmentation followed by disentangled representation learning. However, their design increases model complexity by incorporating style features into the content encoder and does not explicitly enforce domain invariance. In contrast, we perform adversarial domain alignment on both query and image features within the BoQ framework, adding train-time components only and incurring no extra complexity during inference.

\section{Method}
In this section, we first briefly introduce the base architecture of QdaVPR and then present in detail the proposed dual-level adversarial learning framework. Finally, we describe our triplet supervision strategy which leverages query combinations and provide important training details.

\subsection{Base architecture of QdaVPR}
The base architecture of QdaVPR is built on BoQ \cite{BoQ}, which introduces a transformer-based feature aggregation architecture for VPR. Its core idea is the use of a fixed set of learnable global queries, independent of the input image, to probe and aggregate local features extracted by a backbone network. Unlike self-attention mechanisms where queries are dynamically generated from the input, BoQ's static queries provide a consistent and interpretable framework for capturing universal place attributes across varying conditions. Consequently, performing adversarial domain alignment on top of this architecture can effectively yield a domain-agnostic model.

The overall structure of BoQ is depicted in Fig. \ref{fig1} (a) (without the red block). The model processes an input image $I \in {\mathbb{R}^{3 \times H_0 \times W_0}}$, where $(H_0, W_0)$ represents the resolution of the original image, through a backbone (e.g., DINOv2 \cite{dinov2} and a subsequent convolution layer to reduce channels) to obtain a sequence of $N$ local features of dimension $d$: ${{\mathbf{X}}^0} = [{\mathbf{x}}_1^0,{\mathbf{x}}_2^0, \ldots ,{\mathbf{x}}_N^0]$. These features are refined sequentially by a cascade of transformer Encoder units contained in the corresponding BoQ blocks: ${{\mathbf{X}}^l} = {\rm{Encode}}{{\rm{r}}^l}({{\mathbf{X}}^{l - 1}})$, where $l \in \{ 1, 2, \ldots ,L\}$ and $L$ is the total number of BoQ blocks. Each BoQ block also contains a bag of $M$ trainable query vectors: ${{\mathbf{Q}}^l} = [{\mathbf{q}}_1^l,{\mathbf{q}}_2^l, \ldots ,{\mathbf{q}}_M^l]$. Before interacting with the image features, these queries first integrate shared information among themselves via self-attention: ${{\mathbf{Q}}^l} = {\rm{MHA(}}{{\mathbf{Q}}^l},{{\mathbf{Q}}^l},{{\mathbf{Q}}^l}{\mathbf{) + }}{{\mathbf{Q}}^l}$. They then selectively aggregate the refined local features ${{\mathbf{X}}^l}$ through a cross-attention operation: ${{\mathbf{O}}^l} = {\rm{MHA(}}{{\mathbf{Q}}^l},{{\mathbf{X}}^l},{{\mathbf{X}}^l}{\mathbf{)}}$, where MHA is the standard Multi-Head Attention and ${{\mathbf{O}}^l}$ is the output, consisting of $M$ query features. The outputs from all BoQ blocks ${\mathbf{(}}{{\mathbf{O}}^1},{{\mathbf{O}}^2}, \ldots ,{{\mathbf{O}}^L})$ are concatenated and passed through a linear projection layer. The resulting vector is L2-normalized to produce the final global image descriptor. 

In QdaVPR, we extend this architecture by additionally outputting informative representations. This is achieved by splitting the vector from the linear layer into \(N_c\) unnormalized vectors, which are then independently L2-normalized, yielding a set of \(N_c\) query combinations. Each query combination represents a distinct weighted aggregation of the original query features and thus captures complementary aspects of the image content. These query combinations are later used for the triplet supervision described in Section \ref{section_triplet}.

\begin{figure}[!t]
\centering
\includegraphics[width=\linewidth]{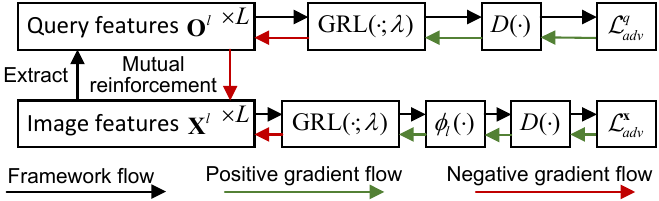}
\caption{Mutual reinforcement in the dual-level adversarial learning framework. Query features improve image features' domain agnosticism through negative gradient flow, and image features reciprocally produce domain-agnostic query features via the framework flow. The symbols are defined in the text.} 
\label{fig2}
\end{figure}

\subsection{Dual-level adversarial learning for domain-agnostic VPR}
To further enhance the domain generalization ability of the BoQ model, we first augment each training image in the GSV-cities dataset with six synthetic domains (fog, rain, snow, wind, night and sun), as illustrated in Fig. \ref{fig1} (b). We then introduce a dual-level adversarial learning framework built upon the base QdaVPR architecture. Note that the adversarial modules are activated only when the training images are from these six augmented domains, while the base architecture remains active throughout training. The motivation for this dual-level design is their mutual reinforcement, as illustrated in Fig. \ref{fig2}: domain-invariant query features encourage the image features to become invariant, and vice versa. This dual-level alignment strengthens the overall domain invariance of the model, ensuring that the final global descriptor is robust to domain shifts.

\textbf{Adversarial learning for the query features.} Each BoQ layer outputs both the extracted query features ${{\mathbf{O}}^l}$ and the image local features ${{\mathbf{X}}^l}$. All query features participate in forming the final global descriptor, so each query feature needs to be domain-agnostic. To achieve this, we introduce adversarial training specifically for the query representations. The query features first pass through a Gradient Reversal Layer (GRL) parameterized by a coefficient $\lambda$ and are then fed into a shared domain discriminator $D(\cdot)$ common to all query features. The discriminator $D$ is implemented as a Multi-Layer Perceptron (MLP) comprising two hidden layers with ReLU activation, formally defined as:
\begin{align}
{{\bf{h}}_1} &= {\mathbf{ReLU}}({{\bf{W}}_1}{\bf{x}} + {{\bf{b}}_1}),\\
{{\bf{h}}_2}& = {\mathbf{ReLU}}({{\bf{W}}_2}{{\bf{h}}_1} + {{\bf{b}}_2}),\\
{\bf{z}} &= {{\bf{W}}_3}{{\bf{h}}_2} + {{\bf{b}}_3},
\end{align}
where \(\mathbf{x}\) is the input feature vector of dimension $d$, \(\mathbf{W}_1, \mathbf{W}_2, \mathbf{W}_3\) and \(\mathbf{b}_1, \mathbf{b}_2, \mathbf{b}_3\) are the learnable weights and biases of the linear layers, and \(\mathbf{z}\) represents the output logits. This discriminator outputs classification logits for the six augmented domains. Its classification ability is optimized by minimizing the cross-entropy loss between these logits and the ground-truth domain labels. The GRL reverses the gradient direction during backpropagation. Consequently, while the loss propagation enhances the discriminator's classification capability, it also drives all query features to become more domain-agnostic by encouraging them to contain less domain-specific information useful for classification. The adversarial loss for the query features is defined by:
\begin{align}
{\cal L}_{adv}^q = \frac{1}{{L \times M}}\sum\limits_{j = 1}^{L \times M} {{\rm{CE}}} \left( {D({\rm{GRL}}({\mathbf{q}_j};\lambda )),{y_d}} \right),
\end{align}
where $\mathbf{q}_j$ represents the $j$-th query feature, $D(\cdot)$ denotes the shared domain discriminator, $\text{GRL}(\cdot; \lambda)$ is the GRL with a coefficient $\lambda$, $y_d$ is the ground-truth domain label, $L \times M$ is the total number of query representations, and CE is the cross-entropy loss function.

\textbf{Adversarial learning for the image feature maps.} Since the query features ${{\mathbf{O}}^l}$ are derived through interaction with the image local features ${{\mathbf{X}}^l}$, if the image local features themselves contain domain-agnostic information, the resulting query features will naturally also be domain-agnostic. Therefore, we similarly introduce adversarial training targeting the image features. Following the same principle, the image features output from each of the $L$ BoQ block, denoted as $\mathbf{X}_l \in \mathbb{R}^{N \times d}$ for block $l$, are first reshaped into 2D feature maps $\mathbf{F}_l = \text{reshape}(\mathbf{X}_l) \in \mathbb{R}^{d \times H \times W}$. They are then passed through the GRL. A dedicated domain feature extractor $\phi_l(\cdot)$ is applied to these reversed features to produce a compact domain representation. For each block $l$, the extractor $\phi_l$ is implemented as a small convolutional network, which processes the input feature map $\mathbf{F}_l$ as follows:
\begin{align}
\mathbf{H}_l^{(1)} &= \text{ReLU}\left( \text{Conv}_{3\times3}(\mathbf{F}_l) \right), \\
\mathbf{H}_l^{(2)} &= \text{AvgPool2d}_{2\times2}(\mathbf{H}_l^{(1)}), \\
\mathbf{H}_l^{(3)} &= \text{ReLU}\left( \text{Conv}_{3\times3}(\mathbf{H}_l^{(2)}) \right), \\
\mathbf{f}_l^{\text{dom}} &= \text{AdaptiveAvgPool2d}_{1\times1}(\mathbf{H}_l^{(3)}),
\end{align}
where \(\text{Conv}_{k\times k}\) denotes a convolutional layer with a \(k \times k\) kernel and padding to maintain spatial dimensions where specified, \(\text{AvgPool2d}_{s\times s}\) is an average pooling layer with kernel size \(s\) and stride \(s\), $\text{AdaptiveAvgPool2d}_{1\times1}$ denotes global average pooling, and \(\mathbf{f}_l^{\text{dom}} \in \mathbb{R}^{d}\) is the final extracted domain feature vector. Crucially, this extracted domain feature \(\mathbf{f}_l^{\text{dom}}\) does not undergo gradient reversal. Thus, it is explicitly encouraged to retain domain-relevant information. The extractor \(\phi_l\) itself learns to capture domain-specific patterns from the spatial feature map.

This domain feature is subsequently fed into the same shared domain discriminator \(D_q(\cdot)\) to predict the domain label. The total adversarial loss for the image features is computed as the average cross-entropy loss across all $L$ blocks::
\begin{align}
\mathcal{L}_{adv}^{\mathbf{x}} = \frac{1}{L} \sum_{l=1}^{L} \text{CE}\left( D\left( \mathbf{f}_l^{\text{dom}} \right), y_d \right).
\end{align}
By applying gradient reversal to the image features \(\mathbf{F}_l\) before domain feature extraction, the optimization simultaneously improves the discriminator's classification performance while driving \(\mathbf{F}_l\) (and consequently the local features \(\mathbf{X}_l\)) to become more domain-agnostic. 

\subsection{Triplet Supervision Based on Query Combinations}
\label{section_triplet}
Standard training pipelines for VPR models utilizing the GSV-cities dataset \cite{gsv-cities} typically adopt its training methodology \cite{ms_loss}. Specifically, for a batch of images sampled from GSV-cities, an online hard mining strategy \(\mathcal{M}\) \cite{ms_loss} is first employed to identify informative triplets:
\begin{align}
(a, p, n) = \mathcal{M}(\mathbf{D}, \mathbf{y}),
\end{align}
where \(\mathbf{D}\) is the set of global descriptors, \(\mathbf{y}\) are the corresponding place labels, \(a\) represents the indices of anchor samples, and \(p\) and \(n\) are the indices of their positive and negative samples within the batch, respectively. Subsequently, a Multi-Similarity (MS) loss \(\mathcal{L}_{MS}\) \cite{ms_loss} guides the optimization of the model's global representation.

However, this supervision is applied solely to the final global descriptor. To inject more VPR-relevant information into the domain-agnostic query features, we leverage the query combinations \(\mathbf{L} \in \mathbb{R}^{B \times N_c \times d}\) (where \(B\) is the batch size and \(d\) the channel dimension) introduced in Fig. \ref{fig1} (a) to construct fine-grained triplet supervision. Our key idea is to treat each query combination independently and identify the most reliable and challenging triplets for loss computation. This design encourages each query combination to become a specialized descriptor that captures complementary visual patterns, thereby enhancing the discriminative power of the final global descriptor. The pseudocode is detailed in Algorithm \ref{alg:triplet-supervision}.

\begin{algorithm}[!t]
\footnotesize
\caption{Query-Combination-Based Triplet Supervision}\label{alg:triplet-supervision}
\begin{algorithmic}
\STATE \textsc{TripletSupervision}$(\mathbf{D}, \mathbf{L}, \mathbf{y}, \alpha, G, H)$
\STATE \textbf{Input:} $\mathbf{D}\in\mathbb{R}^{B\times d}$, $\mathbf{L}\in\mathbb{R}^{B\times N_c\times d}$, $\mathbf{y}\in\mathbb{Z}^B$, $\alpha$, $G$, $H$
\STATE \textbf{Output:} $L_{\text{local}}$
\STATE 
\STATE $( \mathbf{a}, \mathbf{p}, \mathbf{n} ) \gets \textsc{OnlineHardMiner}(\mathbf{D}, \mathbf{y})$   \COMMENT{triplet indices}
Group positives and negatives by anchor: $\mathcal{P}[r]$, $\mathcal{N}[r]$ for unique anchor $r$.
\STATE $L_{\text{total}} \gets 0$, $cnt \gets 0$
\FOR{each unique anchor $r$}
    \STATE $P \gets \mathcal{P}[r]$, $N \gets \mathcal{N}[r]$
    \STATE $\text{sim} \gets \mathbf{D}[r] \cdot \mathbf{D}[N]^\top$ \COMMENT{$[1\times|N|]$}
    \STATE $N_{\text{hard}} \gets \text{top-}G$ indices from $N$ with highest sim
    \STATE $\mathbf{s}_{\text{pos}}, \mathbf{s}_{\text{neg}} \gets \mathbf{0}^{N_c}$
    \FOR{$i = 1$ \TO $N_c$}
        \STATE $\mathbf{f}_r \gets \mathbf{L}[r,i,:]$, $\mathbf{F}_p \gets \mathbf{L}[P,i,:]$, $\mathbf{F}_n \gets \mathbf{L}[N_{\text{hard}},i,:]$
        \STATE $\mathbf{s}_{\text{pos}}[i] \gets \max(\mathbf{f}_r \cdot \mathbf{F}_p^\top)$
        \STATE $\mathbf{s}_{\text{neg}}[i] \gets \max(\mathbf{f}_r \cdot \mathbf{F}_n^\top)$
    \ENDFOR
    \STATE $I_{\text{top}} \gets \text{indices of top-}H$ values in $\mathbf{s}_{\text{pos}}$
    \STATE $L_r \gets \frac{1}{H} \sum_{i \in I_{\text{top}}} \max(0, \alpha - \mathbf{s}_{\text{pos}}[i] + \mathbf{s}_{\text{neg}}[i])$
    \STATE $L_{\text{total}} \gets L_{\text{total}} + L_r$, $cnt \gets cnt + 1$
\ENDFOR
\STATE $L_{\text{local}} \gets \begin{cases} L_{\text{total}} / cnt & \text{if } cnt > 0 \\ 0 & \text{otherwise} \end{cases}$
\RETURN $L_{\text{local}}$
\end{algorithmic}
\end{algorithm}

Specifically, after online hard mining obtains the triplets \((a, p, n)\), for each unique anchor index \(r\) present in the set of anchors \(a\), we process its corresponding query combinations \(\mathbf{L}[r] \in \mathbb{R}^{N_c \times d}\). Due to potential occlusions and long-term structural changes, not every query combination in a positive pair corresponds to genuinely co-visible regions. Therefore, for the \(i\)-th query combination of anchor \(r\), we identify its most similar counterpart among the same query combination index from all its positive samples \(p_r\) as the positive target:
\begin{align}
s^{\text{pos}}_{r,i} = \max_{j \in p_r} \left( \mathbf{L}[r, i] \cdot \mathbf{L}[j, i] \right),
\end{align}
where \(\cdot\) denotes the dot product and $i \in \{ 1, \ldots ,{N_c}\}$. This approach effectively selects the positive sample's query combination that is most similar to the anchor's, ensuring a reliable positive match.

For negative pairs, we first identify a set of hard negatives at the global level. From all negative samples \(n_r\) for anchor \(r\), we select the top-\(G\) samples whose global descriptors are most similar to the anchor's as the hard negative pool \(n^{\text{hard}}_r\). These samples are considered challenging at the query combination level as well. The similarity for the \(i\)-th query combination with these hard negatives is:
\begin{align}
s^{\text{neg}}_{r,i} = \max_{j \in n^{\text{hard}}_r} \left( \mathbf{L}[r, i] \cdot \mathbf{L}[j, i] \right).
\end{align}

Note that not all query combinations are equally discriminative for a given query image. To focus the supervision on the most reliable components, we select the top-\(H\) query combinations (where \(H < N_c\)) with the highest positive similarity scores \(s^{\text{pos}}_{r,i}\). The final localized triplet loss for the entire batch is computed as the average of the triplet losses over these selected combinations across all unique anchors. Let \(N_a\) denote the number of unique anchor indices in the set \(a\). The total local loss \(\mathcal{L}_{\text{local}}\) is then formulated as follows:
\begin{align}
\mathcal{L}_{\text{local}} = \frac{1}{N_a} \sum_{r \in a} \left[ \frac{1}{H} \sum_{i \in \text{Top}H(s^{\text{pos}}_{r,\cdot})} \max\left(0, \,\alpha - s^{\text{pos}}_{r,i} + s^{\text{neg}}_{r,i} \right) \right],
\end{align}
where \(\alpha\) is the margin. This formulation ensures that the supervision is concentrated on the most informative query combinations for each place, thereby encouraging the domain-agnostic query features to encode more distinctive VPR-relevant information.

This fine-grained triplet supervision encourages each domain-agnostic query feature, through its various linear combinations, to encode information that is crucial for distinguishing between places. Consequently, it indirectly enhances the discriminative power of the final global descriptor, leading to more robust VPR performance. The overall training objective combining the global MS loss, the local query combination loss, and the adversarial domain-agnostic losses is:
\begin{align}
\mathcal{L}_{\text{total}} = \mathcal{L}_{MS} + \lambda_{\text{local}} \mathcal{L}_{\text{local}} + \lambda_{\text{adv}}^{q} \mathcal{L}_{adv}^{q} +\lambda_{\text{adv}}^{\mathbf{x}}  \mathcal{L}_{adv}^{\mathbf{x}},
\end{align}
where \(\lambda_{\text{local}}\), \(\lambda_{\text{adv}}^{q}\) and \(\lambda_{\text{adv}}^{\mathbf{x}}\) are balancing hyperparameters.

\section{Experiments}

To comprehensively evaluate the effectiveness and generalization ability of QdaVPR, we conduct extensive experiments on a diverse set of benchmark datasets under various domain shifts. This section details the evaluation datasets, metrics, implementation protocols, and comparative methods, followed by a thorough analysis of experimental results, ablation studies, and visualizations that demonstrate the domain-agnostic properties and superior performance of our approach.

\begin{table}[]
\caption{Overview of the test VPR datasets\label{tab:table1}}
\centering
\footnotesize
\setlength{\tabcolsep}{2.5pt}
\begin{tabular}{cccc}
\toprule[1pt]
Dataset        & Inquiry/Database & Scenery                                                   & Domain Shift  \vspace{0.25em}                                                  \\ \hline
Pitts30k-test \cite{pitts30k}  & 6816/10000       & Urban                                                     & None.                                                            \\ \hline
Tokyo24/7 \cite{tokyo}      & 315/75984        & Urban                                                     & Day/Night.  \\ \hline
MSLS-val \cite{msls}      & 740/18871        & \begin{tabular}[c]{@{}c@{}}Urban,\vspace{-0.25em}\\ Suburban\end{tabular} & \begin{tabular}[c]{@{}c@{}}Day/Night,\vspace{-0.25em}\\ season and weather.\end{tabular}                                                       \\ \hline
AmsterTime \cite{amstertime}    & 1231/1231        & Urban                                                     & \begin{tabular}[c]{@{}c@{}}Long-term,\vspace{-0.25em}\\ modalities.\end{tabular} \\ \hline
SVOX-night \cite{adageo}    & 823/17166        & Urban                                                     & Day/Night.                                                         \\ \hline
SVOX-overcast \cite{adageo} & 872/17166        & Urban                                                     & Weather.                                                         \\ \hline
SVOX-rain \cite{adageo}     & 937/17166        & Urban                                                     & Weather.                                                         \\ \hline
SVOX-snow \cite{adageo}     & 870/17166        & Urban                                                     & Weather.                                                         \\ \hline
SVOX-sun \cite{adageo}      & 854/17166        & Urban                                                     & Weather.
                                 \\ \hline
\begin{tabular}[c]{@{}c@{}}Nordland \cite{nordland},\vspace{-0.25em}\\ Summer/Winter \end{tabular}    & 27592/27592        & \begin{tabular}[c]{@{}c@{}}Natural,\vspace{-0.25em}\\ Suburban\end{tabular}                                                     & Season.                                                         \\ \hline
\begin{tabular}[c]{@{}c@{}}Nordland \cite{nordland},\vspace{-0.25em}\\ Summer/Spring \end{tabular}     & 27592/27592        & \begin{tabular}[c]{@{}c@{}}Natural,\vspace{-0.25em}\\ Suburban\end{tabular}                                                     & Season.                                                         \\ \hline
\begin{tabular}[c]{@{}c@{}}Nordland \cite{nordland},\vspace{-0.25em}\\ Summer/Fall \end{tabular}    & 27592/27592        & \begin{tabular}[c]{@{}c@{}}Natural,\vspace{-0.25em}\\ Suburban\end{tabular}                                                     & Season. 
\\ \Xhline{1pt}
\end{tabular}
\end{table}

\subsection{Evaluation datasets}
To evaluate the generalization ability of our QdaVPR, we conduct experiments on the following test datasets. Their key
properties are listed in Table. \ref{tab:table1}

The Pitts30k-test \cite{pitts30k} contains 6,816 inquiries and 10,000 Google Street View database images, depicting urban scenes without domain shift.

Tokyo24/7 \cite{tokyo} includes 315 cellphone inquiries and 75,984 Street View database images, introducing day/night variation (database images are daytime only).

MSLS-val \cite{msls} offers 740 inquiries and 18,871 database images from urban and suburban areas, featuring domain shifts due to seasons, weather, and illumination.

AmsterTime \cite{amstertime} provides 1,231 pairs of urban images, each pairing a grayscale historical inquery with an RGB reference, challenging modality and long-term variations.

SVOX \cite{adageo} leverages a Street View database of Oxford and inquiries from the Oxford RobotCar dataset \cite{oxford}, which includes various weather and illumination conditions, such as overcast, rainy, sunny, snowy and nighttime.

Nordland \cite{nordland} provides images captured from a fixed front-facing viewpoint on a train across four seasons. The scenes primarily show suburban and natural landscapes, with frame-level correspondence provided as ground truth. Following previous works \cite{eigenplaces}, we extract images at 1FPS, using summer as the reference season and spring, fall, or winter as the inquiry seasons. Note that Nordland in the following experiments refers to the Nordland-Summer/Winter test set.

\subsection{Evaluation metrics}
We follow the evaluation metric widely used in previous research \cite{SciceVPR,salad,edtformer,djist,ImAge,selavpr,structvpr++,FoL}, where Recall@N is measured on the VPR datasets. Recall@N is defined as the percentage of the inquiry images for which at least one of the first N predictions is from the same place. For Pitts30k-test, Tokyo24/7, MSLS-val and SVOX datasets with GPS labels, a predicted database image is considered to be from the same place as an inquiry if their distance is within 25 meters. AmsterTime consists of images pairs, where only the counterpart of an inquiry image in the database images comes from the same place as the inquiry. For the Nordland dataset, we apply a tolerance of 10 frames, meaning each inquiry image is matched to a set of 20 reference images from the database sequence. In the rest of the paper, R@N refers to Recall@N.

\subsection{Implementation and training details}
QdaVPR is trained on the original GSV-cities \cite{gsv-cities} and its 6 augmented varied-domain datasets using PyTorch on a single 4090 GPU with a batch size of 640. Each batch consists of 160 places, each containing 4 images, and each image is equally and randomly drawn from the 7 GSV-cities datasets to ensure that the total number of training images remains the same as in the original GSV-cities. The backbone of QdaVPR is based on DINOv2-B \cite{dinov2} followed by a $3\times3$ convolution to reduce and fuse the channels, which is similar to BoQ \cite{BoQ}. QdaVPR extracts features from the last 4 layers of DINOv2-B, which are then concatenated and fed into the $3\times3$ convolution. The feature channel dimension in each DINOv2-B layer is 768, resulting in a concatenated feature dimensionality of 3072, which is subsequently reduced to 384 through the $3\times3$ convolution. The weights of the DINOv2-B in QdaVPR are initialized from the pretrained weights of the DINOv2-B in BoQ and are frozen during the training process. The number of BoQ layers is set to 2, with each layer containing 64 learnable queries, and the total of 128 query representations are reduced to 32 via a linear layer. Each query representation has a dimensionality of 384, so the output dimensionality of the global descriptor is 12288, and PCA is applied to reduce the dimensionality. The hidden dimensionality of the shared domain discriminator is 512, where the first linear layer expands the feature dimension from 384 to 512, the second linear layer maintains the dimension at 512, and the last linear layer maps the feature dimension to 6 logits. Each domain feature extraction module preserves the feature channel dimension, meaning that every $3\times3$ convolution outputs features with a channel dimension of 384. The hyperparameter for the GRL is $\lambda=-1$. The loss weights are \(\lambda_{\text{local}}=0.01\), \(\lambda_{\text{adv}}^{q}=0.05\) and \(\lambda_{\text{adv}}^{\mathbf{x}}=0.05\). For the Algorithm \ref{alg:triplet-supervision}, the hyperparameters are $\alpha=0.05$, $G=10$ and $H=8$. The training images are resized to $280 \times 280$, whereas the testing images are resized to $322 \times 322$ (unless specified), following the same configuration as BoQ. The model is optimized using the AdamW \cite{adamw} optimizer with a weight decay of 0.001 and a linear warmup for 10 epochs, and the total number of training epochs is 40. The initial learning rate is set to 0.0003 and is multiplied by 0.1 every 10 epochs, and basic data augmentation techniques \cite{randomaug} are applied during training. The QdaVPR checkpoint attaining the best R@1 on MSLS-val \cite{msls} is chosen for evaluation.

\begin{table*}[]
\caption{R@N comparison to SOTA methods on benchmark datasets. The best results are in \textbf{bold} and the second best \underline{underlined}. 
\label{tab:table2}}
\centering
\footnotesize
\setlength{\tabcolsep}{2.5pt}
\begin{tabular}{l|c|ccc|cccccc|ccc|ccc}
\Xhline{1pt}
\multirow{2}{*}{Method} & \multirow{2}{*}{Image size} & \multicolumn{3}{c|}{Nordland \cite{nordland}}   & \multicolumn{3}{c|}{Tokyo24/7 \cite{tokyo}}                                     & \multicolumn{3}{c|}{MSLS-val \cite{msls}}                 & \multicolumn{3}{c|}{AmsterTime \cite{amstertime}}               & \multicolumn{3}{c}{Pitts30k-test \cite{pitts30k}}                  \\ \cline{3-17} 
                        &                             & R@1           & R@5           & R@10          & R@1           & R@5           & \multicolumn{1}{c|}{R@10}          & R@1           & R@5           & R@10          & R@1           & R@5           & R@10          & R@1           & R@5           & R@10          \\ \hline
SALAD \cite{salad}, CVPR2024         & \multirow{7}{*}{$224 \times 224$}    & 79.8          & 90.2          & 93.4          & 91.7          & 96.5          & \multicolumn{1}{c|}{96.5}          & 91.1          & 95.5          & 96.4          & 53.0          & 74.4          & 79.5          & 91.7          & 95.7          & 97.0          \\
BoQ \cite{BoQ}, CVPR2024           &                             & 83.9          & 92.4          & 95.0          & 94.3          & 97.5          & \multicolumn{1}{c|}{97.8}          & 90.9          & 95.7          & 96.4          & 53.6          & 72.5          & 76.9          & 93.0          & 96.4          & 97.3          \\
EDTFormer \cite{edtformer}, TCSVT2025    &                             & 80.9          & 91.2          & 94.2          & \textbf{96.5} & \underline{97.8}    & \multicolumn{1}{c|}{\underline{98.1}}    & \underline{91.9}    & \textbf{96.1} & \textbf{96.6} & \underline{58.9}    & \underline{80.3}    & \underline{84.7}    & 92.8          & 96.8          & 97.8          \\
ImAge \cite{ImAge}, NIPS2025         &                             & 88.2          & 95.5          & 97.1          & 92.1          & 96.2          & \multicolumn{1}{c|}{97.8}          & 90.7          & 95.9          & \textbf{96.6} & 56.3          & 76.1          & 81.6          & \underline{93.2}    & 96.8          & 97.8          \\
CerfeVPR \cite{CerfeVPR}, CMC2025       &                             & \textbf{92.5} & \textbf{96.8} & \textbf{97.8} & \multicolumn{3}{c}{\textbackslash{}}                               & 87.7          & 94.2          & 95.3          & \multicolumn{3}{c|}{\textbackslash{}}         & \textbf{94.9} & \textbf{97.5} & \underline{98.0}    \\
SciceVPR \cite{SciceVPR}, NEUCOM2026    &                             & 73.2          & 86.0          & 90.2          & \underline{95.6}    & \textbf{98.4} & \multicolumn{1}{c|}{\textbf{98.4}} & 89.5          & 95.5          & \underline{96.5}    & \textbf{59.1} & \textbf{81.9} & \textbf{85.9} & 92.9          & \underline{97.0}    & \textbf{98.1} \\
QdaVPR(ours)            &                             & \underline{89.3}    & \underline{95.7}    & \underline{97.2}    & 95.2          & 97.5          & \multicolumn{1}{c|}{\textbf{98.4}} & \textbf{92.0} & \textbf{96.1} & \underline{96.5}    & 57.9          & 77.7          & 82.2          & 92.7          & 96.4          & 97.3          \\ \hline
SALAD \cite{salad}, CVPR2024         & \multirow{5}{*}{$322 \times 322$}    & 85.1          & 93.1          & 95.3          & 94.3          & 97.1          & \multicolumn{1}{c|}{97.5}          & 91.9          & 96.4          & \underline{96.9}    & 57.8          & 78.5          & 83.6          & 92.3          & 96.2          & 97.4          \\
BoQ \cite{BoQ}, CVPR2024           &                             & 88.8          & 95.2          & 96.9          & 96.5          & \underline{98.1}    & \multicolumn{1}{c|}{\textbf{99.0}} & \underline{93.0}    & 96.5          & 96.8          & 59.5          & 79.2          & 82.1          & \underline{93.6}    & 96.8          & 97.6          \\
EDTFormer \cite{edtformer}, TCSVT2025    &                             & 88.5          & 95.2          & 97.0          & \underline{97.1}    & \textbf{98.4} & \multicolumn{1}{c|}{\underline{98.4}}    & 92.4          & \underline{96.8}    & \textbf{97.2} & \textbf{64.9} & \textbf{85.1} & \textbf{88.8} & \underline{93.6}    & \underline{97.0}    & \underline{97.9}    \\
ImAge \cite{ImAge}, NIPS2025         &                             & \underline{92.7}    & \underline{97.4}    & \underline{98.4}    & 95.9          & \underline{98.1}    & \multicolumn{1}{c|}{\underline{98.4}}    & 92.8          & \textbf{96.9} & \textbf{97.2} & \underline{61.1}    & 80.7          & 85.3          & \textbf{94.0} & \textbf{97.2} & \textbf{98.0} \\
QdaVPR(ours)            &                             & \textbf{93.5} & \textbf{97.7} & \textbf{98.6} & \textbf{97.5} & 97.8          & \multicolumn{1}{c|}{\textbf{99.0}} & \textbf{93.6} & 95.9          & 96.2          & \underline{61.1}    & \underline{81.8}    & \underline{86.2}    & 93.3          & 96.6          & 97.3          \\ \Xhline{1pt}
\end{tabular}
\end{table*}

% \usepackage{multirow}
% \usepackage[normalem]{ulem}
\begin{table*}[]
\caption{R@N comparison to SOTA methods on the challenging benchmark datasets under varied domain conditions. Inference images are resized to $322 \times 322$. The best results are in \textbf{bold} and the second best \underline{underlined}. 
\label{tab:table3}}
\centering
\footnotesize
\setlength{\tabcolsep}{1.5pt}
\begin{tabular}{c|cccccc|ccccccccccccccc}
\Xhline{1pt}
\multirow{3}{*}{Method} & \multicolumn{6}{c|}{Nordland \cite{nordland}}                                                                                      & \multicolumn{15}{c}{SVOX \cite{adageo}}                                                                                                                                                                                                                                                                                                         \\ \cline{2-22} 
                        & \multicolumn{3}{c|}{Summer/Spring}                                 & \multicolumn{3}{c|}{Summer/Fall}              & \multicolumn{3}{c|}{Rain}                                          & \multicolumn{3}{c|}{Sun}                                           & \multicolumn{3}{c|}{Snow}                                          & \multicolumn{3}{c|}{Night}                                         & \multicolumn{3}{c}{Overcast}                  \\ \cline{2-22} 
                        & R@1           & R@5           & \multicolumn{1}{c|}{R@10}          & R@1           & R@5           & R@10          & R@1           & R@5           & \multicolumn{1}{c|}{R@10}          & R@1           & R@5           & \multicolumn{1}{c|}{R@10}          & R@1           & R@5           & \multicolumn{1}{c|}{R@10}          & R@1           & R@5           & \multicolumn{1}{c|}{R@10}          & R@1           & R@5           & R@10          \\ \hline
SALAD \cite{salad}                  & 96.2          & 98.3          & \multicolumn{1}{c|}{98.9}          & 99.4          & \underline{99.7}    & \underline{99.7}    & \underline{98.6}    & \underline{99.7}    & \multicolumn{1}{c|}{\textbf{99.8}} & 97.3          & 99.1          & \multicolumn{1}{c|}{\underline{99.5}}    & 98.6          & 99.5          & \multicolumn{1}{c|}{\underline{99.7}}    & 95.5          & \underline{99.3}    & \multicolumn{1}{c|}{99.4}          & \textbf{98.4} & \underline{99.3}    & 99.3          \\
BoQ \cite{BoQ}                    & 97.3          & 99.1          & \multicolumn{1}{c|}{99.4}          & \underline{99.5}    & \textbf{99.8} & \textbf{99.8} & 98.5          & 99.6          & \multicolumn{1}{c|}{\underline{99.7}}    & 97.4          & \underline{99.4}    & \multicolumn{1}{c|}{99.4}          & \textbf{99.3} & \textbf{99.8} & \multicolumn{1}{c|}{\textbf{99.8}} & 97.2          & 98.9          & \multicolumn{1}{c|}{99.4}          & \underline{98.3}    & 99.2          & \underline{99.4}    \\
EDTFormer \cite{edtformer}              & 97.3          & 98.9          & \multicolumn{1}{c|}{99.4}          & \textbf{99.6} & \textbf{99.8} & \textbf{99.8} & 98.3          & \textbf{99.8} & \multicolumn{1}{c|}{\textbf{99.8}} & \underline{98.4}    & \textbf{99.5} & \multicolumn{1}{c|}{\textbf{99.8}} & 98.5          & \underline{99.7}    & \multicolumn{1}{c|}{\textbf{99.8}} & 95.7          & 98.8          & \multicolumn{1}{c|}{\underline{99.5}}    & 98.1          & 99.2          & \underline{99.4}    \\
ImAge \cite{ImAge}                  & \underline{97.6}    & \underline{99.2}    & \multicolumn{1}{c|}{\underline{99.5}}    & 99.4          & \underline{99.7}    & \textbf{99.8} & 98.0          & 99.6          & \multicolumn{1}{c|}{99.6}          & 97.4          & \underline{99.4}    & \multicolumn{1}{c|}{\underline{99.5}}    & \underline{98.7}    & 99.5          & \multicolumn{1}{c|}{\underline{99.7}}    & \underline{97.3}    & 99.1          & \multicolumn{1}{c|}{\textbf{99.6}} & 97.6          & 99.1          & 99.2          \\
QdaVPR(ours)            & \textbf{98.3} & \textbf{99.3} & \multicolumn{1}{c|}{\textbf{99.6}} & \textbf{99.6} & \textbf{99.8} & \textbf{99.8} & \textbf{98.9} & \textbf{99.8} & \multicolumn{1}{c|}{\textbf{99.8}} & \textbf{98.7} & \underline{99.4}    & \multicolumn{1}{c|}{99.4}          & \textbf{99.3} & \underline{99.7}    & \multicolumn{1}{c|}{\textbf{99.8}} & \textbf{97.8} & \textbf{99.6} & \multicolumn{1}{c|}{\textbf{99.6}} & \underline{98.3}    & \textbf{99.5} & \textbf{99.5} \\ \Xhline{1pt}
\end{tabular}
\end{table*}

\subsection{Compared methods}
To fairly compare VPR models with different backbones and training datasets, we consider only existing methods \cite{BoQ,salad,edtformer,SciceVPR,ImAge,CerfeVPR} that adopt DINOv2-B \cite{dinov2} as the backbone and are trained on GSV-cities \cite{gsv-cities} or its derivatives (i.e., synthetic images generated by CycleGAN \cite{cyclegan}). GSV-cities inherently encompasses diverse domain shifts (i.e., illumination, weather, and seasonal changes), and models trained on it implicitly acquire domain-agnostic information through metric learning \cite{ms_loss}. In contrast, CerfeVPR explicitly enriches this process by generating summer and winter domain images via CycleGAN, which are fed into the network alongside the original GSV-cities images. It then utilizes a transformer to extract domain-agnostic representations across these domains. Our method, by comparison, constructs a training set composed of the original GSV-cities and six augmented variants, while maintaining the total number of training images equivalent to that of the original GSV-cities through careful sampling. We achieve domain-agnostic VPR via dual-level adversarial learning, and further boost the discriminative power of the output features through triplet supervision based on query combinations. Note that all compared models have an output dimensionality of 4096, and we compare various inference image sizes, except SciceVPR and CerfeVPR, which inherently only accept an input image size of $224 \times 224$.

% \usepackage{multirow}
% \usepackage[normalem]{ulem}
% \useunder\underline{ine}\underline{}{}

\subsection{Experimental results}
This section analyzes the results of QdaVPR compared with other SOTA VPR models on challenging benchmark datasets in Table \ref{tab:table1}. Since SciceVPR and CerfeVPR are constrained to input image size of $224 \times 224$, we first compare our method with them at this resolution, then evaluate QdaVPR at $322 \times 322$ alongside other leading approaches. A qualitative comparison of retrieval results is detailed in the \textbf{supplemental material}.

As shown in Table \ref{tab:table2} with input image set to $224 \times 224$, QdaVPR achieves competitive performance across various domain shifts. Specifically, it ranks in the top three for Recall@1, Recall@5, and Recall@10 on all datasets with domain variations, with the pitts30k-test set (which has no domain shift) being the only exception. In contrast, SciceVPR (SOTA on AmsterTime) degrades considerably on the Nordland dataset, which features summer-winter seasonal variations; CerfeVPR, while achieving SOTA on Nordland through season-specific augmentations, underperforms on MSLS-val where weather and illumination vary.

In Tables \ref{tab:table2} and \ref{tab:table3} with input image set to $322 \times 322$, QdaVPR achieves the best Recall@1 on most test scenarios, whereas the compared models struggle to handle the various domain shifts present in each dataset. For instance, EDTFormer exhibits poor performance on the Nordland-Summer/Winter test set, despite achieving SOTA results on AmsterTime. ImAge attains SOTA on the Pitts30k-test set, but its Recall@1 on datasets with domain shifts is consistently lower than that of QdaVPR. These results demonstrate the SOTA performance of QdaVPR regardless of domain shifts. 

\subsection{Ablation studies}
We first select the hyperparameters of the loss weights, which also demonstrate the effectiveness of our dual-level adversarial learning and query-combination-based triplet supervision. Then, we further determine the optimal hyperparameters for Algorithm \ref{alg:triplet-supervision}. Finally, we show how QdaVPR, compared to the original BoQ model, achieves superior place recognition performance even when the global descriptor dimension is reduced. All the compared QdaVPR variants produce 4096-dimensional global descriptors with inference image size $322 \times 322$, unless specified.

% \usepackage[normalem]{ulem}
% \usepackage[normalem]{ulem}
\begin{table}[]
\caption{Ablation on loss weights. 'Original' and 'Augmented' refer to the GSV-cities dataset used for training. The best R@1 results are in \textbf{bold} and the second best \underline{underlined}. 
\label{tab:table4}}
\centering
\footnotesize
\setlength{\tabcolsep}{2.5pt}
\begin{tabular}{ccc|c|c|c|c}
\Xhline{1pt}
$\lambda_{\text{adv}}^{q}$    & $\lambda_{\text{adv}}^{\mathbf{x}}$    & $\lambda_{\text{local}}$ & Train set                   & MSLS-val      & Nordland      & Tokyo24/7     \\ \hline
0    & 0    & 0     & original                    & 92.7          & 88.0          & \underline{97.1}    \\ \hline
0    & 0    & 0     & \multirow{18}{*}{augmented} & 92.7          & 89.3          & 96.8          \\
0    & 0.05 & 0     &                             & 93.0          & 93.5          & \underline{97.1}    \\
0.01 & 0.05 & 0     &                             & 92.3          & 92.9          & 96.2          \\
0.1  & 0.05 & 0     &                             & 92.2          & 89.0          & 96.2          \\
0.5  & 0.05 & 0     &                             & 92.6          & 92.8          & 96.5          \\
0.05 & 0    & 0     &                             & 92.6          & 91.2          & 96.2          \\
0.05 & 0.01 & 0     &                             & 92.4          & 92.7          & \underline{97.1}    \\
0.05 & 0.1  & 0     &                             & 92.2          & \underline{93.6}    & 96.5          \\
0.05 & 0.5  & 0     &                             & 92.3          & 91.7          & \underline{97.1}    \\
0.05 & 0.05 & 0     &                             & \underline{93.5}    & 92.7          & 96.5          \\
0.05 & 0.05 & 0.005 &                             & 93.0          & 87.8          & 96.5          \\
\textbf{0.05} & \textbf{0.05} & \textbf{0.01}  &                             & \textbf{93.6} & 93.5          & \textbf{97.5} \\
0.05 & 0.05 & 0.05  &                             & 93.0          & 91.7          & \textbf{97.5} \\
0.05 & 0.05 & 0.1   &                             & 93.0          & 90.6          & 96.2          \\
0.05 & 0.05 & 0.5   &                             & \underline{93.5}    & 88.2          & 96.5          \\
0    & 0.05 & 0.01  &                             & 92.3          & \textbf{93.8} & 96.2          \\
0.05 & 0    & 0.01  &                             & 93.4          & 89.7          & \underline{97.1}    \\
0    & 0    & 0.01  &                             & 93.1          & 92.0          & 96.2          \\ \Xhline{1pt}
\end{tabular}
\end{table}

\textbf{Ablation on loss weights.} We adopt a two-step strategy to determine the optimal loss weights. First, we identify the optimal weights for the dual-level adversarial learning components, $\lambda_{\text{adv}}^{q}$ and $\lambda_{\text{adv}}^{\mathbf{x}}$. Subsequently, with these fixed, we find the optimal weight for the triplet supervision based on query combination, $\lambda_{\text{local}}$. All weights are selected based on performance on the MSLS-val set, following the model selection protocol of BoQ.

As shown in Table \ref{tab:table4}, employing the augmented GSV-cities dataset for training, even without any adversarial learning ($\lambda_{\text{adv}}^{q}=0$, $\lambda_{\text{adv}}^{\mathbf{x}}=0$), provides a performance baseline. Introducing either form of adversarial learning individually improves domain invariance. Applying adversarial learning to the query features alone ($\lambda_{\text{adv}}^{q}=0.05$) boosts the performance on Nordland from 89.3\% to 91.2\%. A more pronounced effect is observed when it is applied to the image feature maps ($\lambda_{\text{adv}}^{\mathbf{x}}=0.05$), which increases Nordland performance to 93.5\%. The first stage of mining for the dual-level adversarial weights reveals that applying both strategies simultaneously ($\lambda_{\text{adv}}^{q}=0.05$, $\lambda_{\text{adv}}^{\mathbf{x}}=0.05$) yields the highest performance on the challenging MSLS-val dataset, achieving 93.5\%.

In the second stage, we tune the local loss weight $\lambda_{\text{local}}$ with the adversarial weights fixed at $\lambda_{\text{adv}}^{q}=0.05$ and $\lambda_{\text{adv}}^{\mathbf{x}}=0.05$. As shown in Table \ref{tab:table4}, setting $\lambda_{\text{local}}=0.01$ further improves the model's performance on MSLS-val to 93.6\%. This configuration also achieves the best result on the Tokyo24/7 dataset while maintaining a competitive score on Nordland.

Finally, to verify the contribution of each adversarial component, we fix $\lambda_{\text{local}}=0.01$ and ablate them one by one, as shown in the last three rows of Table \ref{tab:table4}. Removing either the query feature level ($\lambda_{\text{adv}}^{q}=0$) or the image feature level ($\lambda_{\text{adv}}^{\mathbf{x}}=0$) adversarial learning leads to a noticeable performance drop on at least one of the target domains. This confirms that both levels of the dual adversarial learning are essential for achieving the most robust and balanced domain-invariant global descriptors, while the addition of an appropriate triplet supervision based on query combinations further enhances their discriminative power.

% \usepackage[normalem]{ulem}
\begin{table}[]
\caption{Ablation on hyperparameters of the query-combination-based triplet supervision. The best R@1 results are in \textbf{bold} and the second best \underline{underlined}. 
\label{tab:table5}}
\centering
\footnotesize
\setlength{\tabcolsep}{2.5pt}
\begin{tabular}{ccc|c|c|c}
\Xhline{1pt}
$\alpha$    & $G$  & $H$  & MSLS-val      & Nordland      & Tokyo24/7     \\ \hline
0.05 & 10 & 4  & 92.8          & 93.3          & \underline{97.1}    \\
\textbf{0.05} & \textbf{10} & \textbf{8}  & \textbf{93.6} & 93.5          & \textbf{97.5} \\
0.05 & 10 & 12 & 93.0          & \textbf{94.7} & 96.5          \\
0.05 & 10 & 16 & 92.4          & 92.9          & 96.8          \\
0.05 & 10 & 20 & 93.1          & 92.8          & \underline{97.1}    \\
0.05 & 10 & 24 & 93.1          & 91.0          & 96.5          \\
0.05 & 10 & 28 & 93.2          & 92.8          & 96.5          \\
0.05 & 10 & 32 & 92.2          & 93.1          & 96.5          \\
0.01 & 10 & 8  & 92.2          & 89.4          & 96.8          \\
0.1  & 10 & 8  & 92.4          & 91.8          & 96.8          \\
0.05 & 5  & 8  & \underline{93.4}    & 89.9          & 96.5          \\
0.05 & 15 & 8  & 93.1          & 93.2          & 96.5          \\
0.05 & 20 & 8  & 91.8          & \underline{93.8}    & 96.5          \\ \Xhline{1pt}
\end{tabular}
\end{table}

% \usepackage{multirow}
\begin{table}[]
\caption{Ablation on descriptor dimensionality with the best R@1 results in \textbf{bold}. 
\label{tab:table6}}
\centering
\footnotesize
\setlength{\tabcolsep}{2pt}
\begin{tabular}{c|c|c|c|c|c}
\Xhline{1pt}
Method                        & Dimension & MSLS-val      & Nordland      & Tokyo24/7     & AmsterTime    \\ \hline
\multirow{4}{*}{BoQ \cite{BoQ}}          & 12288     & \textbf{93.8} & 90.6          & \textbf{96.5} & \textbf{63.0} \\
                              & 4096      & 93.0          & 88.8          & \textbf{96.5} & 59.5          \\
                              & 2048      & 92.2          & 85.4          & 95.2          & 54.8          \\
                              & 512       & 85.8          & 68.4          & 88.6          & 34.9          \\ \hline
\multirow{4}{*}{QdaVPR(ours)} & 12288     & \textbf{93.6} & \textbf{94.1} & 97.1          & \textbf{64.0} \\
                              & 4096      & \textbf{93.6} & 93.5          & \textbf{97.5} & 61.1          \\
                              & 2048      & 92.6          & 92.2          & 96.5          & 57.0          \\
                              & 512       & 90.5          & 82.5          & 94.0          & 45.7          \\ \Xhline{1pt}
\end{tabular}
\end{table}

\textbf{Ablation on hyperparameters of the triplet supervision based on query combinations.} Table \ref{tab:table5} summarizes the results. Based on the R@1 on MSLS-val, we select the optimal configuration. Deviating $\alpha$ from 0.05 consistently degrades performance. Reducing the number of hard negatives $G$ (from 10 to 5) severely hurts Nordland accuracy, while increasing $G$ (to 15 or 20) mainly affects MSLS-val. Varying $H$ has a relatively modest impact on MSLS-val, though larger fluctuations are observed on Nordland. Therefore, we set $\alpha=0.05$, $G=10$, and $H=8$, as this combination achieves the best R@1 on MSLS-val.

\textbf{Ablation on descriptor dimensionality.} As reported in Table \ref{tab:table6}, our QdaVPR exhibits significantly smaller performance drops than BoQ when the descriptor dimension is reduced. Except for the 12288-dim case on MSLS-val (where BoQ leads by 0.2\%), our model achieves superior R@1 on all datasets and dimensions, underscoring its effectiveness with low-dimensional global descriptors.

\begin{figure*}[!t]
\centering
\includegraphics[width=0.8\linewidth]{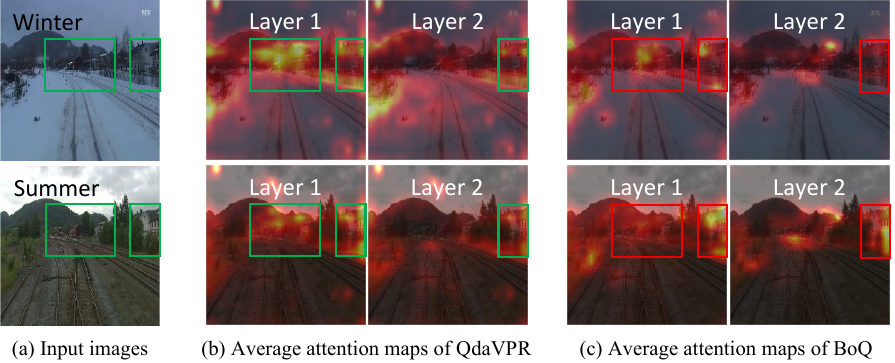}
\caption{Average attention map visualization for (b) QdaVPR and (c) BoQ \cite{BoQ}. In the input images (a), green blocks indicate regions containing buildings. In (b) and (c), green blocks denote consistently high attention to the buildings shown in (a) across different weather conditions, whereas red blocks indicate high attention under only one specific weather condition and low attention under the other.}
\label{fig3}
\end{figure*}

\subsection{Visualization}
To qualitatively illustrate why QdaVPR can be regarded as a domain-agnostic VPR model, we visualize the average cross-attention weights between the input images and the learned queries in each BoQ layer for both QdaVPR and the original BoQ \cite{BoQ} model. As shown in Fig. \ref{fig3} (b) and (c), each BoQ layer in QdaVPR maintains consistently high attention to the same buildings across different weather conditions (highlighted in green in Fig. \ref{fig3} (a)), whereas the attention in the original BoQ model varies with weather conditions. This observation explains the superior performance of QdaVPR on the Nordland-Summer/Winter benchmark, where it outperforms BoQ by over 3\% in Table \ref{tab:table6}, and also demonstrates why QdaVPR is domain-agnostic: it pays nearly equal attention to the same buildings regardless of environmental variations. 

\section{Conclusion}
In this paper, we propose QdaVPR, a novel query-based domain-agnostic model for VPR. By introducing a dual-level adversarial learning framework that operates on both query representations and the underlying image features, QdaVPR effectively learns domain-invariant global descriptors. Furthermore, a query-combination-based triplet supervision enhances the discriminative power of these descriptors, enabling robust place recognition across diverse environmental conditions. Extensive experiments on multiple benchmarks with significant domain shift demonstrate that QdaVPR achieves SOTA performance while introducing no additional computational overhead during inference. Visualization analyses further confirm that the model consistently attends to salient structural landmarks (e.g., buildings) regardless of domain changes. However, we observe that QdaVPR yields relatively lower performance on datasets without domain shift, indicating a trade-off between domain generalization and peak performance on homogeneous data. Future work will aim to develop a domain-agnostic VPR model that maintains high accuracy even on scenarios without domain variations.

\bibliographystyle{IEEEtran}

\bibliography{references.bib}

\vfill

\end{document}